\documentclass[runningheads]{llncs}
\usepackage{cite}
\usepackage{amsmath,amssymb,amsfonts}
\usepackage{times}
\usepackage{caption}
\usepackage{graphicx}
\usepackage{textcomp}
\usepackage{xcolor}
\usepackage{multirow}
\usepackage{subcaption}
\usepackage{algorithm}
\usepackage[noend]{algpseudocode}
\usepackage{hyperref}       
\usepackage{url}            
\def\BibTeX{{\rm B\kern-.05em{\sc i\kern-.025em b}\kern-.08em
    T\kern-.1667em\lower.7ex\hbox{E}\kern-.125emX}}

\DeclareMathOperator*{\argmin}{arg\,min}

\begin{document}
\title{Robust Influence-based Training Methods for Noisy Brain MRI}

\author{Minh-Hao Van\orcidID{0000-0001-7342-6801} \and
Alycia N. Carey\orcidID{0000-0002-3587-4088} \and
Xintao Wu~\thanks{Corresponding Author} \orcidID{0000-0002-2823-3063}}

\authorrunning{M. Van et al.}
\institute{University of Arkansas, Fayetteville AR 72701, USA \\
\email{\{haovan,ancarey,xintaowu\}@uark.edu}}

\maketitle 

\begin{abstract}
Correctly classifying brain tumors is imperative to the prompt and accurate treatment of a patient. While several classification algorithms based on classical image processing or deep learning methods have been proposed to rapidly classify tumors in MR images, most assume the unrealistic setting of noise-free training data. In this work, we study a difficult but realistic setting of training a deep learning model on noisy MR images to classify brain tumors. We propose two training methods that are robust to noisy MRI training data, Influence-based Sample Reweighing (ISR) and Influence-based Sample Perturbation (ISP), which are based on influence functions from robust statistics. Using the influence functions, in ISR, we adaptively reweigh training examples according to how helpful/harmful they are to the training process, while in ISP, we craft and inject helpful perturbation proportional to the influence score. Both ISR and ISP harden the classification model against noisy training data without significantly affecting the generalization ability of the model on test data. We conduct empirical evaluations over a common brain tumor dataset and compare ISR and ISP to three baselines. Our empirical results show that ISR and ISP can efficiently train deep learning models robust against noisy training data.

\end{abstract}

\section{Introduction}
Accurate brain tumor detection and diagnosis is a crucial step in providing appropriate medical treatment and improving the prognosis of individuals with brain cancer and other brain ailments. However, the human brain is complex, and it can be difficult to correctly determine both the type and stage of a brain tumor. Often, magnetic resonance imaging (MRI) is utilized to create 3-D scans of the brain, which a specialist can use to plan a course of treatment based on the determined tumor type and stage. Unfortunately, analyzing MR images is a non-trivial task, and relying on human perception alone can result in misdiagnosis and delayed treatment. More and more, domain experts are relying on computer software that can analyze large-scale MR images in rapid time. Many image processing methods \cite{john2012brain,zulpe2012glcm,cheng2015enhanced,bosch2006modeling,avni2010x,ayadi2022brain} have been proposed to classify brain tumors based on type by extracting useful features from an MR image to feed into the classification algorithm. More recently, techniques from deep learning have also been used to improve the performance of medical image classification systems \cite{abiwinanda2019brain,deepak2019brain,afshar2018brain}.

Despite the advances in classifying MR images, many of the proposed works assume an unrealistic setting of noise-free data. Obtaining clean and high-quality MR images unperturbed by noise is challenging and time-consuming. Noise can be generated by a variety of different sources during the process of an MR scan (e.g., the patient moving, the sampling procedure used, or hardware issues), and in most cases collecting noisy MRI data is unavoidable. When the MR images are noisy, the performance ability of a classifier significantly decreases in terms of accuracy.  Therefore, it is important to consider the setting where noisy data is assumed. 
In this work, we simulate the challenging but realistic scenario of the training and test dataset containing noisy MR images. Specifically, we propose two training methods, named Influence-based Sample Reweighing (ISR) and Influence-based Sample Perturbation (ISP), using the influence function from robust statistics to improve the generalization ability of a classification model trained on noisy brain tumor data.

\section{Related Work}
\label{sec:related-work}
Improving the generalization ability of models trained to classify medical images has been a significant area of research over the past several years. 
While the field is broad, many of the proposed works utilize classical image processing techniques to extract features from the medical images to use in the training of a classification model. 
For example, the wavelet transform \cite{john2012brain} and the Gray Level Co-occurrence Matrix (GLCM) \cite{zulpe2012glcm, john2012brain} are two commonly used methods from image processing that can be used to construct intensity and texture features of the MR images. However, processing techniques from outside classical image processing have been proposed as well. For example, Bag-of-Words (BoW) method, which was originally proposed in the text-mining field to build a representation vector for each word in a document, has been shown to be capable of representing the complex features of a medical image \cite{cheng2015enhanced, bosch2006modeling, avni2010x}. Some approaches combining traditional and non-traditional image processing methods have also been proposed. For instance, \cite{ayadi2022brain} combines classical feature extractors (HoG and DSUFT) and a Support Vector Machine (SVM) to improve the performance and efficiency of the classifier in a multi-class setting. Additionally, \cite{cheng2015enhanced} showed that when tumor regions are available, they can serve as a region of interest (ROI) and be augmented using image dilation to help boost the accuracy of BoW, the intensity histogram, and the GLCM.

With the growing popularity of deep learning, many works have proposed to adopt deep neural networks (DNNs) in the medical imaging field. In \cite{abiwinanda2019brain}, the authors evaluated the effectiveness of different Convolutional Neural Network (CNN) architectures on a brain tumor classification task, and they found that a simple CNN architecture can achieve good performance without any prior knowledge of the segmented tumor regions. Further, transfer learning has been proven to be an effective method in helping a model to be more generalizable when there is limited training data across several different settings, and \cite{deepak2019brain} showed that transfer learning is also useful to boost generalizability when working with medical images. \cite{afshar2018brain} ventured beyond classic DNNs and showed that using more advanced architectures like Capsule Networks \cite{hinton2011transforming} can help to improve the accuracy of brain tumor classification, and they additionally explored why CapsNet tends to over-fit the data more than classic CNN models do.

Most of the works discussed above are proposed (and shown to be successful) 
based on the assumption that both the training and test datasets are free of noise. However, noisy data, in general, is unavoidable due to randomness and error during data collection. There are many robust training methods proposed to alleviate the potential harm of noisy data to a classification model \cite{goodfellow2014explaining,wong2020fast}. 
Specifically in the medical imaging domain, \cite{xie2022effective,wu2020classification} showed that adversarial training methods are effective when adapted to medical image analysis. Further \cite{krull2019noise2void,lehtinen2018noise2noise} showed that image denoising methods can be utilized to remove noise and reconstruct clean medical images without requiring prior knowledge of clean data.

\section{Preliminaries}
\label{sec:prelims}
\subsection{Influence Function on Single Validation Point}
\label{sec:prelims_single}

Influence functions, first proposed in the field of robust statistics \cite{cook1982residuals}, 
have recently been found useful for explaining machine learning models. In \cite{koh2017understanding}, the authors proposed a method for estimating the influence that a single training point $z=(x,y)$ has on both the parameters of a machine learning model and the loss of a single test point. Here $x \in X$ represents a feature in the domain of features $X$ and $y \in Y$ represents a label in the domain of labels $Y$. Let $f_\theta$ be a classification model parameterized by $\theta$, $\hat{\theta}$ be the optimal parameters of $f$, and  $D_{trn}/D_{val}/D_{tst}$ be the training/validation/test datasets. Let $l(\cdot, \theta)$ represent the loss and 
$L(D_{\ast}, \theta) = \dfrac{1}{|D_{\ast}|}\sum_{z_i \in D_{\ast}}l(z_i, \theta)$ be the empirical loss function to be minimized, where $D_{\ast}$ denotes ``any dataset''.
To see how a training point $z$ affects the model's parameters, we write the empirical risk minimization as:
\begin{equation}
    \label{eq:erm}
    \hat{\theta}_{\epsilon, z}=\argmin_{\theta \in \Theta}\dfrac{1}{|D_{trn}|}\sum_{z_i \in D_{trn}}l(z_i,\theta)+\epsilon l(z, \theta)
\end{equation}
In Eq. \ref{eq:erm}, we simulate the removal of $z$ from the training set by upweighting it by a small weight $\epsilon$ (usually on the order of $-\frac{1}{n}$ where $n$ is the number of training points). Rather than calculating Eq. \ref{eq:erm} through training, \cite{koh2017understanding} shows that the influence of training point $z$ on the model parameters can be estimated as:
\begin{equation}
\label{eq:up-param}
    \mathcal{I}_{up, param}(z) = \dfrac{d\hat{\theta}_{\epsilon,z}}{d\epsilon}\Big|_{\epsilon=0}=-H^{-1}_{\hat{\theta}}\nabla_{\theta}l(z,\hat{\theta})
\end{equation}
and the resulting change in parameters can be calculated as $\hat{\theta}_{\epsilon,z}\approx \hat{\theta}-\frac{1}{n}\mathcal{I}_{up,params}(z)$. In addition to showing the effect a training point $z$ has on the model parameters, \cite{koh2017understanding} extends Eq. \ref{eq:up-param} to calculate the influence that a training point $z$ has on a test point $z_{test}$.
\begin{equation}
    \mathcal{I}_{up, loss}(z, z_{test}) = -\nabla_{\theta}l(z_{test}, \hat{\theta})^{\top}H_{\hat{\theta}}^{-1}\nabla_{\theta}l(z, \hat{\theta}) 
    \label{eq:ss_up_loss}
\end{equation}

Both Eq. \ref{eq:up-param} and Eq. \ref{eq:ss_up_loss} estimate the effect of a training point $z$ being completely removed from the training dataset. However, in \cite{koh2017understanding}, the authors additionally show that influence functions can be used to estimate the effect of perturbing $z=(x,y)$ to $\hat{z}=(x+\delta,y)$ on the model parameters. The empirical risk minimization can be written as:
\begin{equation}
    \label{ref:approx-param}
    \hat{\theta}_{\epsilon, \hat{z}, -z} = \argmin_{\theta \in \Theta} \dfrac{1}{|D_{trn}|}\sum_{z_i\in D_{trn}}l(z_i, \theta)-\epsilon l(\hat{z}, \theta)+\epsilon l(z,\theta)
\end{equation}
and Eq. \ref{ref:approx-param} can be estimated as $\hat{\theta}_{\epsilon, \hat{z}, -z} \approx \hat{\theta} -\frac{1}{n}\mathcal{I}_{pert,param}(z)$ where:
\begin{equation}
\label{eq:pert-param}
    \begin{aligned}
    \mathcal{I}_{pert,param}(z) &= -H^{-1}_{\hat{\theta}}\left(\nabla_\theta l(\hat{z}, \hat{\theta})-\nabla_\theta l(z, \hat{\theta})\right)
\end{aligned}
\end{equation}
As in the case with Eq. \ref{eq:up-param}, \cite{koh2017understanding} extends Eq. \ref{eq:pert-param} to show how perturbing $z\to \hat{z}$ would affect the loss of a test point $z_{test}$:
\begin{equation}
    \mathcal{I}_{pert, loss}(z, z_{test}) = -\nabla_{\theta}l(z_{test}, \hat{\theta})^{\top}H_{\hat{\theta}}^{-1}\nabla_x\nabla_{\theta}l(z, \hat{\theta})
    \label{eq:ss_pert_loss}
\end{equation}
The main difference between Eq. \ref{eq:ss_up_loss} and Eq. \ref{eq:ss_pert_loss} is that in Eq. \ref{eq:ss_pert_loss}, the gradient of $\nabla_{\theta}L(z, \hat{\theta})$ w.r.t $x$ is additionally calculated. This additional gradient computation captures how changing $z$ along each dimension of $x$ affects the loss of a test point. 

\subsection{Influence Function on Validation Group} 
\label{sec:prelims_group}

It is essential to note that Eqs. \ref{eq:ss_up_loss} and \ref{eq:ss_pert_loss} consider the influence that a single training point has on a single test point. 
However, calculating the influence score with respect to a single test point may not produce a good estimation when the training data is noisy. Further, it is computationally expensive to calculate the influence score for each pair of training and test points individually because each calculation requires the inverse Hessian matrix to be computed (either directly or via estimation). Therefore, we extend the influence functions of Eqs. \ref{eq:ss_up_loss} and \ref{eq:ss_pert_loss} to estimate the impact that a single training point has on a group of test (or validation) points. Since influence is additive \cite{koh2017understanding}, we can extend both equations to consider how the loss of a group of validation points changes when $z$ is either removed or perturbed:
\begin{equation}\small
    \mathcal{I}_{up, loss}(z, D_{val}) = -\nabla_{\theta}L\left(D_{val}, \hat{\theta}\right)^{\top}H_{\hat{\theta}}^{-1}\nabla_{\theta}l\left(z, \hat{\theta}\right)
    \label{eq:sg_up_loss}
\end{equation}
\begin{equation}\small
    \mathcal{I}_{pert, loss}(z, D_{val}) = -\nabla_{\theta}L\left(D_{val}, \hat{\theta}\right)^{\top}H_{\hat{\theta}}^{-1}\nabla_x\nabla_{\theta}l\left(z, \hat{\theta}\right)
    \label{eq:sg_pert_loss}
\end{equation}

With deep learning models, the top layers generally serve as classification layers while the lower layers work as feature extractors. Therefore, instead of computing the influence score over all the deep learning model's layers, we can simply use the top-most layers. 
However, computing the inverse Hessian matrix ($H_{\hat{\theta}}^{-1}$) is still computationally intensive even if we only consider the top-most layers. We leverage the inverse Hessian-vector product (IHVP) method to approximate $H_{\theta}^{-1}\nabla_{\theta}L(D_{val}, \hat{\theta})$ in Eq. \ref{eq:sg_up_loss} and Eq. \ref{eq:sg_pert_loss} using the Linear time Stochastic Second-Order Algorithm (LiSSA) \cite{agarwal2017second}.

\section{Influence-based Sample Reweighing}
\label{sec:ISR}
We propose Influence-based Sample Reweighing (ISR) -- a training procedure robust to noisy MR image data that uses influence scores to reweigh the training examples. 

\subsection{Framework}
\label{sec:ISR_fw}

Most robust training approaches either manipulate the entire training dataset, or a portion of it, to train a model robust to noisy training points. One popular approach is to introduce a weight for each training point, which forces the model to pay more attention to important samples. We formulate our ISR robust training procedure by modifying the empirical loss function introduced in Section \ref{sec:prelims} to incorporate sample weights. Let $w \in \mathbb{R}^{|D_{trn}|}$ be the vector of weights for the training data where weight $w_i$ corresponds to training point $z_i$. The per-sample weighted loss function can be defined as:
\begin{equation}
    L_w(D_{trn}, \theta, w) = \dfrac{1}{|D_{trn}|} \sum_{z_i \in D_{trn}} w_i l(z_i,\theta)
    \label{eq:w_loss}
\end{equation}
The loss function $L_w(D_{trn}, \theta, w)$ in Eq. \ref{eq:w_loss} is a generalized formulation of an empirical loss function. 
Recall from Eq. \ref{eq:sg_up_loss} that the influence score can tell us the effect that a training point has on the validation loss. For ISR, we construct the weight vector $w$ based on the influence score of each training point. Since $\mathcal{I}_{up, loss}(z, D_{val})$ reflects the influence that training point $z$ has on the validation loss, we can determine the weight of each training sample based on if they have a negative (helpful) or positive (harmful) score. We introduce our ISR in Algorithm \ref{alg:train_isr}.
\begin{algorithm}[ht]
    \begin{algorithmic}[1]
        \renewcommand{\algorithmicrequire}{ \textbf{Input:}}
        \renewcommand{\algorithmicensure}{ \textbf{Output:}}
        \Require Training data $D_{trn}$, validation data $D_{val}$, train epochs $T$, pre-train epochs $T_{pre}$, learning rate $\eta$, maximum weight $m$
        \Ensure Trained model $\hat{\theta}$
        \State Initialize $\theta^0$
        \For{$t=1\dots T_{pre}$}
            \State $\theta^t \leftarrow \theta^{t} - \eta\nabla L(D_{trn},\theta^t) $
        \EndFor
        \State Initialize $w$ as an array of size $|D_{trn}|$ \Comment{Calculating sample weights in lines 4-7} 
        \For{$i = 1 \dots |D_{trn}|$}
            \State $w_i \leftarrow \mathcal{I}_{up,loss}(z_i, D_{val})$ using Eq. \ref{eq:sg_up_loss}
        \EndFor
        \State Scale $w$ to range $[1,m]$ using MinMax scaler 
        \For{$t=T_{pre}+1\dots T$}
            \State $\theta^t \leftarrow \theta^{t} - \eta \nabla L_{w}(D_{trn},\theta^t, w) $
        \EndFor
    \end{algorithmic}
    \caption{Training with Influence-based Sample Reweighing}
    \label{alg:train_isr}
\end{algorithm}

In Algorithm \ref{alg:train_isr}, the model is randomly initialized with $\theta^0$ and then trained for $T_{pre}$ epochs in lines 1-3. Normally, calculating influence scores requires a fully trained model. However, we find that allowing the model to complete several pre-training steps sufficiently warms the $\theta$ parameters to generate good influence scores. In lines 4-7, ISR calculates the weights for training samples. Finally, we continue the training procedure using the per-sample weighted loss function for the remaining epochs in lines 8-9. 

\subsection{Calculating Sample Weights}
Lines 4-7 of Algorithm \ref{alg:train_isr} shows how to calculate the weights $w_i$'s of the training points based on their influence score. After initializing $w$ in line 4, we go through each training sample, estimate its influence score using Eq. \ref{eq:sg_up_loss}, and assign the score to the vector $w$ in lines 5-6. Since the influence score values are usually close to zero and can be either negative or positive, the loss function will be deactivated if we directly use the influence scores as weights. Therefore, in line 7, we scale the values in $w$ to be in the range from 1 to $m$ using the MinMax scaler. Specifically, samples with negative (positive) influences should have weights close to 1 ($m$). By doing so, helpful samples keep their original effect on the model, while harmful samples are penalized and weighted higher such that the model will pay more attention to them. The weighted loss in the ISR training procedure is therefore bounded based on the empirical loss:
\begin{equation}
    L(D_{\ast}, \theta) \leq L_w(D_{\ast}, \theta, w) \leq mL(D_{\ast}, \theta)
\end{equation}
where equality only holds when $m=1$. When $m$ approaches 1, the weighted loss approximates the empirical loss, and the addition of weights has no impact on the learning of the model. On the other hand, when $m \gg 1 $, it takes longer for the model to converge as the loss is sufficiently larger. For this reason, in our experiments, we choose $m=2$.

\section{Influence-based Sample Perturbation}
\label{sec:ISP}
We counteract the noise in a training image by adding perturbation proportional to the image's influence score -- a process we call Influence-based Sample Perturbation (ISP). 

\subsection{Framework}
\label{sec:framework}
ISP works as follows: (1) ISP selects a subset of training points $D_s$ which have the most impact on the model by calculating the influence of each training point on the model loss and selecting those with the highest influence score; and (2) ISP generates (helpful) perturbation for every example in $D_s$ based on the influence score to fortify the model against noisy training examples. More specifically, we construct a subset $D_s\subset D_{trn}$ by choosing the most influential examples in $D_{trn}$ and let $D_u = D_{trn}\setminus D_s$. Let $\delta_i$ be the influence-based perturbation for training point $z_i = (x_i, y_i)$ and let $\hat{z}_i=(x_i+\delta_i, y_i)$ be the influence-based perturbed version of $z_i$.
We define the ISP training set as $\hat{D}_{trn}=\hat{D}_s\cup D_u$, where $\hat{D}_s$ is the set $D_s$ \textit{after} influence-based perturbation is added. Under this setting, we define the empirical loss function of the robust model as $L(\hat{D}_{trn}, \theta) = L(\hat{D}_{s}, \theta) + L(D_u, \theta)$. We define our objective as:
\begin{equation}
    \min_{\delta \in \Delta} \ L(D_{val}, \theta_{\delta}) \text{ s.t. } \theta_{\delta} = \argmin_{\theta \in \Theta} L(\hat{D}_{trn}, \theta) \label{eq:noise_train}
\end{equation}
The na\"{i}ve approach to the above problem would be to train/optimize several different models with different perturbation values $\delta_i$, which is intractable. By using influence functions, we avoid the requirement of costly retraining. Specifically, we use the influence function of Eq. \ref{eq:sg_up_loss} to estimate the change in the model's loss on $D_{val}$ when a particular training point is perturbed to train a model robust to noisy input data efficiently. We fully present the robust training procedure ISP in Algorithm \ref{alg:hint}. 

\begin{algorithm}[ht]
    \begin{algorithmic}[1]
        \renewcommand{\algorithmicrequire}{ \textbf{Input:}}
        \renewcommand{\algorithmicensure}{ \textbf{Output:}}
        \Require Training data $D_{trn}$, validation data $D_{val}$, train epochs $T$, pre-train epochs $T_{pre}$, scaling factor $\gamma$, ratio of selected examples $r$, learning rate $\eta$
        \Ensure Trained model $\hat{\theta}$
        \State Initialize $\theta^0$
        \For{$t=1\dots T_{pre}$}
            \State $\theta^t \leftarrow \theta^{t} - \eta\nabla L(D_{trn},\theta^t) $
        \EndFor
        \State $D_{s} \leftarrow \emptyset$ \Comment{Selecting influential samples in lines 4-9}
        \For{$z_i \in D_{trn}$}
            \State Compute $-\mathcal{I}_{up,loss}(z_i, D_{val})$ using Eq. \ref{eq:sg_up_loss}
        \EndFor
        \State Sort data points in $D_{trn}$ in ascending order of influence scores
        \State Assign the first $\lceil r\times |D_{trn}|\rceil$ examples of $D_{trn}$ to $D_{s}$
        \State $D_{u} \leftarrow D_{trn} \setminus D_{s}$
        \State $\hat{D}_s \leftarrow D_s$ \Comment{Adding influence-based perturbation in lines 10-15}
        \For{$z_i \in D_s$}
            \State $\delta_i \leftarrow -\gamma\mathcal{I}_{pert, loss}(\hat{z}_i, D_{val})$
            \State $\hat{x}_i \leftarrow \text{Clip}(x_i+\delta_i)$
            \State $\hat{z}_i \leftarrow (\hat{x}_i, y_i)$
            \State Update new $\hat{z}_i$ in $\hat{D}_s$
        \EndFor
        \State $\hat{D}_{trn} \leftarrow \hat{D}_{s} \cup D_u$   
        \For{$t=T_{pre}+1\dots T$}
            \State $\theta^t \leftarrow \theta^{t} - \eta \nabla L(\hat{D}_{trn},\theta^t) $
        \EndFor
    \end{algorithmic}
    \caption{Training with Influence-based Sample Perturbation}
    \label{alg:hint}
\end{algorithm}
Similar to ISR, Algorithm \ref{alg:hint} begins by pre-training the model parameterized by $\theta$ for $T_{pre}$ epochs (lines 2-3). The parameters in early epochs of training can be unstable, and it is crucial to avoid adding perturbation to the training examples based on unstable parameters. In lines 4-9, we select the most influential training points $D_s$ (and thereby the points not in $D_s$, $D_u$) (discussed in Section \ref{sec:secinf}). In lines 10-16, we first add perturbation to selected training examples (lines 10-15), and then update $\hat{D}_{trn}$ (line 16). The complete details of how the influence-based perturbation is generated will be given in Section \ref{sec:add_noise}. The model then undergoes additional training on the updated training set $\hat{D}_{trn}$ (lines 17 and 18) to construct a model robust to noisy examples.

\subsection{Selecting Influential Samples}
\label{sec:secinf}
The data distribution of a (clean) training dataset, and the same training dataset with added noisy samples, are not the same. In other words, the noisy samples shift the data distribution and, therefore, will shift the model's decision boundary away from the one generated when the model is trained on the clean data. Further, training samples that are highly affected by noise will tend to stand apart in the loss space \cite{yang2022not}. Intuitively, we can increase the generalization ability of the model by focusing on the training samples which have a harmful impact on the model. 

Recall from Section \ref{sec:prelims} that the influence function tells how the model parameters (or loss) would change if a data point $z$ was removed from the training set. We can utilize this idea to select the training examples which cause the increase in the model's loss when they are included in the training phase. We present our subset selection method based on influence scores. For each sample in the training data, we calculate the influence score using the upweighting approach of Eq. \ref{eq:sg_up_loss} in lines 5-6. We then build the subset $D_s$ (and $D_u$) by selecting the $\lceil r\times |D_{trn}|\rceil$ examples, which have the highest impact in lines 7-9.

\subsection{Adding Influence-based Perturbation}
\label{sec:add_noise}
In Eq. \ref{eq:sg_pert_loss}, $\mathcal{I}_{pert, loss}(z, D_{val})$ tells us the contribution of each input pixel to the loss of the whole validation set. While the gradient of the loss of the validation set, $\nabla_{\theta}L(D_{val}, \hat{\theta})$, provides information on how the trained model performs on unseen validation data, the term $\nabla_x\nabla_{\theta}l(z, \hat{\theta})$ tells how each input pixel contributes to the loss. Pixels with either significantly positive or negative influence scores highlight where the model's attention is focused. Since the influence score represents the change in loss, a pixel that has a positive (negative) score will cause an increase (decrease) in the loss. Therefore, by crafting perturbation in the opposite direction of the influence score, i.e., $\delta=-\mathcal{I}_{pert, loss}(z, D_{val})$, we can perturb the original image in a way that further strengthens helpful pixels and weakens harmful pixels.
\begin{figure}[ht]
    \centering
    \includegraphics[width=.50\textwidth]{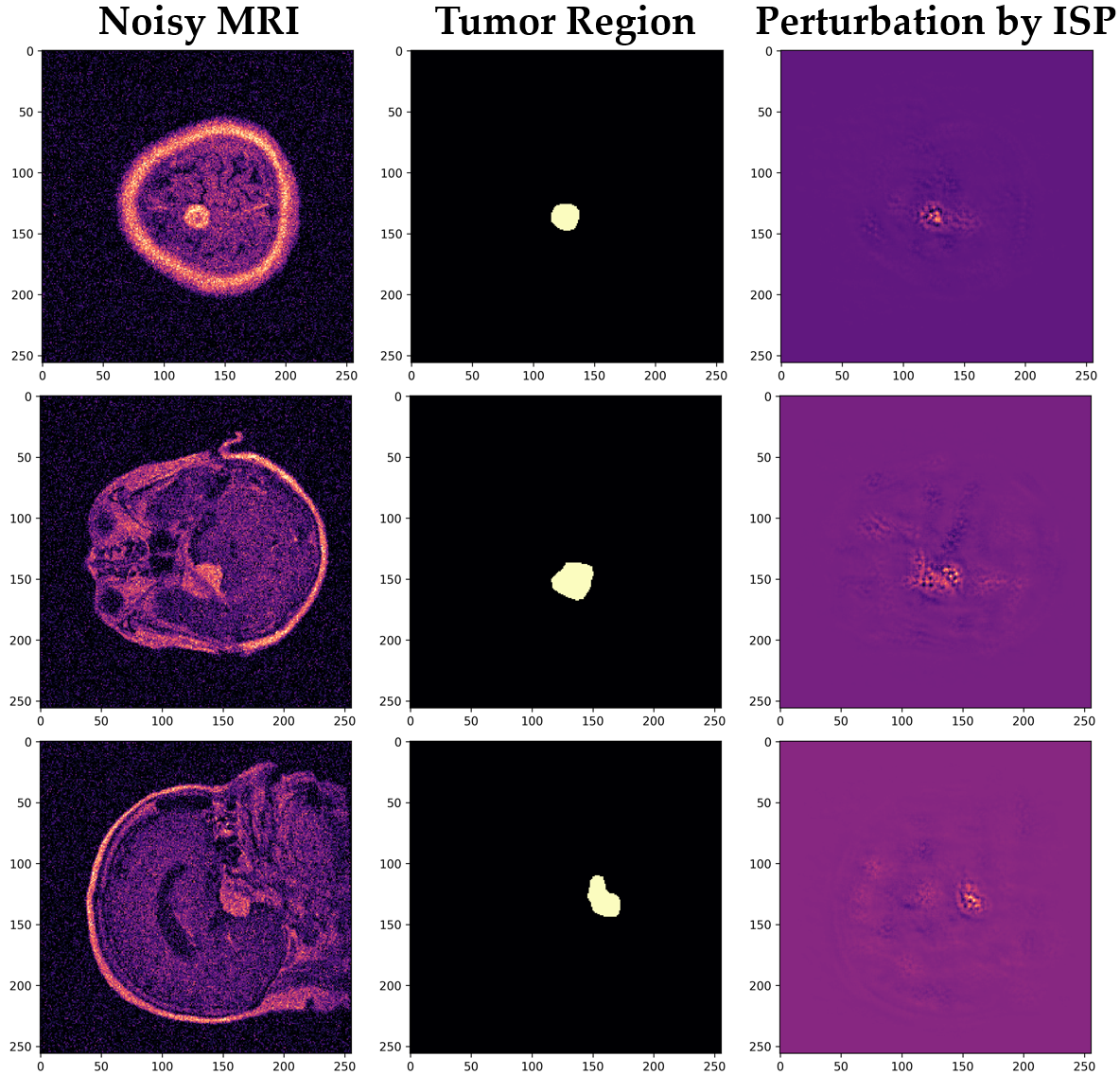}
    \caption{Visualization of perturbation by our ISP.}
    \label{fig:example_isp}
\end{figure}

Let $\delta_i \in \Delta$ be the influence-based perturbation corresponding to training input $x_i$, i.e., $\hat{x}_i=x_i+\delta_i$ (line 12). Adding $\delta_i$ to $x_i$ could potentially result in $\hat{x}_i$ being a non-feasible image. Therefore, we clip the perturbed image $\hat{x}_i$ (line 13) before setting the perturbed training point as $\hat{z}_i = (\hat{x}_i, y)$ (line 14). Finally, we add the newly perturbed example $\hat{z}_i$ to $\hat{D}_s$ in line 15. 
In Fig. \ref{fig:example_isp}, we visualize training MR images and influence-based perturbations to show how our ISP can make the model more robust by adding friendly perturbation. The first column shows the noisy MR image. The second column is the information on the tumor region provided by the dataset. The last column is the influence-based perturbation generated by our ISP. Looking at the second and last columns, we can see that the brighter region in the perturbation aligns with the tumor region. ISP leads the attention of the classifier to the critical regions of an image by adding more perturbation to those areas.

\section{Experiments}
\label{sec:experiments}
\subsection{Evaluation Setup}
\label{sec:setup}
\subsubsection{Dataset and Model}
\label{sec:dataset}
For all experiments conducted, we use the Brain Tumor Dataset \cite{Cheng2017}. The dataset contains 3,064 brain T1-weighted CE-MRI slices with a size of $512\times 512$ from 233 patients. There are three kinds of tumors: 708 meningiomas, 1426 gliomas, and 930 pituitary images. We allocate 75\% of the patients to be in the training set, 5\% to be in the validation set, and 20\% to be in the test set. 
We use a Convolutional Neural Network (CNN) as our classification model. The architecture of the CNN model consists of five convolutional layers (32, 32, 64, 64, and 64). The last layer is a fully connected layer. We use Batch Normalization, ReLU activation, and Max Pooling after each convolutional layer to improve the performance and efficiency of the model. Data augmentation (e.g., random horizontal flipping) is incorporated to increase the diversity of the training set. The model is trained for 40 epochs using the Adam optimizer with a learning rate of 0.01 and batch size of 32.

\subsubsection{Simulating Noisy Images}
We craft the noisy brain tumor MR images by adding random noise from either: 1) the Gaussian distribution centered at $\mu=0$ with a standard deviation of $\sigma=32$, or 2) the Rician distribution located at $\nu=1$ with a scale of $\sigma=16$. In our evaluation, we aim to simulate a difficult operating scenario where the majority of the labeled data is noisy. Consequently, we add noise to every image in the training data and leave the validation data clean. For the test data, we add random noise to a subset of the original images where the size of the subset is determined by a parameter $\rho_{test}$. We set $\rho_{test} = \{0, 0.5\}$ to show the effectiveness of our methods.
Our evaluation setting is realistic and mirrors the reality that only a small number of collected MR images are noise-free, while the majority have some degree of noise. 

\subsection{Proposed Methods and Baselines}
We compare our proposed methods with the following three baselines: 

\noindent\textbf{Na\"ive Training:} In this setting, the CNN model is trained on the original brain tumor dataset, and no noise correction or modified training procedure is applied. 

\noindent\textbf{Adversarial Training (AT):} This method hardens the base CNN model by crafting adversarial examples and optimizing the model over the generated adversarial data. We use FGSM with uniform initialization \cite{wong2020fast} to generate the adversarial perturbations. I.e., $\delta = \alpha\cdot \text{sign}(\nabla_x l(f_{\theta}(x),y))$, where in our experiments we set $\alpha=0.03$ and $\epsilon=0.1$. 

\noindent\textbf{Noise2Void (N2V):} 
N2V aims to reconstruct clean images from the given noisy data with a U-Net \cite{ronneberger2015u} backbone architecture and uses a blind-spot network to train the denoiser. 
We train the N2V model using the default settings from \cite{krull2019noise2void}, and we train a CNN model on the denoised images using settings defined in Section \ref{sec:setup}. In the results, we denote a combined method of N2V denoising and CNN training as N2V-Training to make it distinguishable from N2V, which only involves image denoising.

\noindent\textbf{ISR and ISP:} To efficiently compute the influence scores, we only use the weights from the fully connected layer of the CNN model (see Section \ref{sec:prelims_group} for more details). We choose $T_{pre}$ as 10, $r=0.1$, $\gamma=0.01$ and use the CNN architecture.

\noindent\textbf{Metrics:} 
For each experiment, we report the average and standard deviation of the test accuracy over five trials. We use a GPU Tesla V100 (32GB RAM) and CPU Xeon 6258R 2.7 GHz to conduct all experiments.

\subsection{Results}
\subsubsection{Intervened Training Data by Methods}
\label{sec:intervened_data}
\begin{figure}[t]
    \centering   
    \includegraphics[width=.90\textwidth]{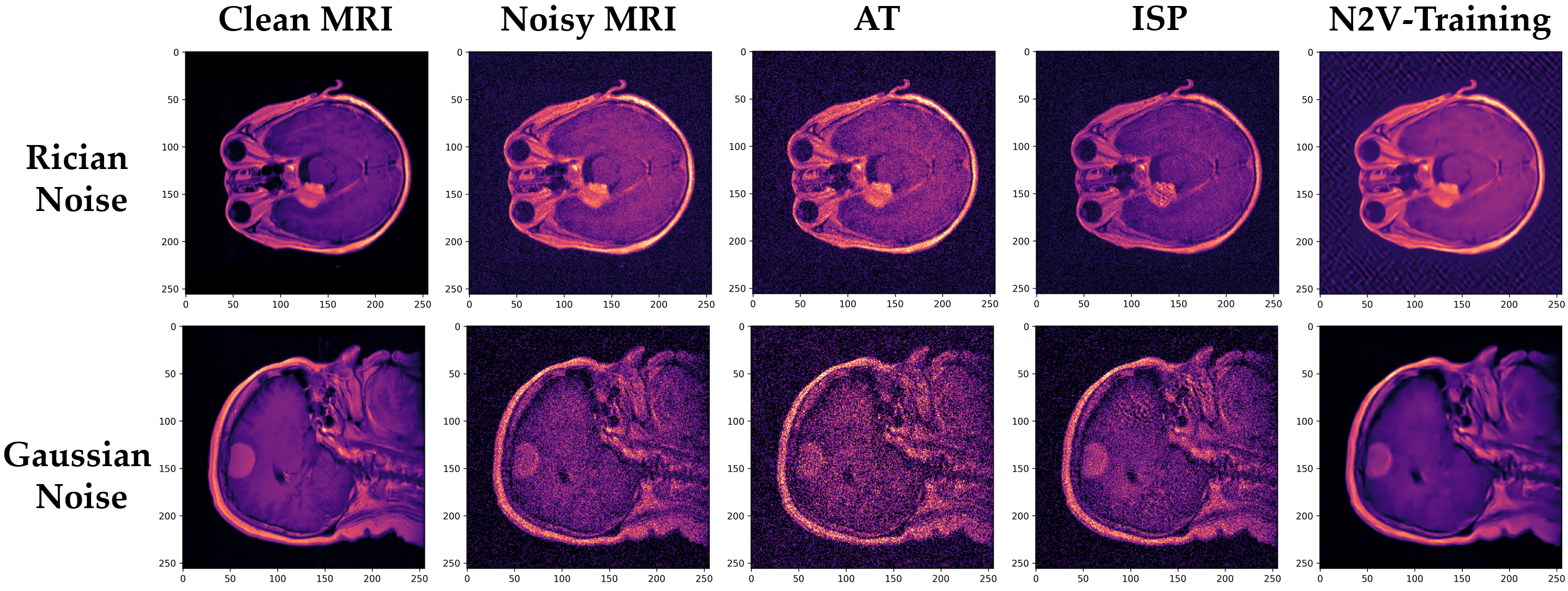}
    \caption{Training samples crafted by each training method.}
    \label{fig:noise_training}
\end{figure}

While ISR puts the model's attention on important samples via reweighing the training loss, training methods like AT, our ISP, and N2V-Training intervene on the noisy training images. In Fig. \ref{fig:noise_training}, we illustrate the training images generated by the AT, ISP, and N2V-Training. Note that since our ISR only reweighs the images, the noisy training image is not perturbed, and hence we do not provide a column for ISR in Fig. \ref{fig:noise_training}.

AT and ISP craft additional perturbation to inject into the training images to increase the generalization ability of the model. By comparing images perturbed by AT and our ISP (third and fourth columns) to clean and noisy images (first and second columns), we can see that ISP alleviates the harmful noise and makes the generated images look closer to the clean ones. While ISP can significantly reduce the noise in the background (darker background), AT, on the contrary, makes the value of background noise larger (brighter background). In other words, our ISP is more successful than AT in reverting harmful noise. ISP also makes the tumor regions, which are the key regions of the classification task, more detailed. For example, in the first row, the tumor region of the ISP image has a clearer boundary and is more prominent than the surrounding regions. This observation is clearer when looking at the example of Gaussian Noise on the second row. 

N2V-Training is able to generate images with less noise than AT and ISP because it denoises the training images before training the classifier. However, generated images can lose important fine-grained information in the tumor image. Further, when considering Rician noise, N2V-Training generates an image where the brain itself is free from noise, but the background noise becomes stronger.
\subsubsection{Overall Performance}
\begin{table}[t]
    \centering
    \caption{Test accuracy (\%) of training methods on noisy Brain Tumor Dataset}
    \label{tab:test_rho}
    \renewcommand{\arraystretch}{1.25}
    \setlength{\tabcolsep}{5pt}
    \begin{tabular}{c|c|c|c|c}
        \hline
        \multirow{2}{*}{Method} & \multicolumn{2}{c|}{$\rho_{test}=0$} & \multicolumn{2}{c}{$\rho_{test}=0.5$} \\ \cline{2-5}
        & Gaussian Noise & Rician Noise & Gaussian Noise & Rician Noise \\
        \hline
        Na\"ive & 87.77 $\pm$ 0.84 & 86.50 $\pm$ 3.38 & 88.89 $\pm$ 0.88 & 88.44 $\pm$ 2.43 \\ \hline
        AT & 88.38 $\pm$ 1.47 & 86.85 $\pm$ 1.62 & 88.95 $\pm$ 0.94 & 89.04 $\pm$ 0.54 \\ \hline
        N2V-Training & 88.12 $\pm$ 1.69 & 83.92 $\pm$ 2.52 & 89.08 $\pm$ 1.70 & 86.78 $\pm$ 1.79 \\ \hline
        ISR (Our) & 88.79 $\pm$ 1.75 & 88.12 $\pm$ 2.28 & 89.17 $\pm$ 1.09 & 89.49 $\pm$ 1.82 \\
        ISP (Our) & \textbf{89.52 $\pm$ 2.61} & \textbf{89.49 $\pm$ 1.45} & \textbf{90.73 $\pm$ 1.58} & \textbf{90.54 $\pm$ 1.19} \\ \hline
    \end{tabular}
\end{table}
Table \ref{tab:test_rho} shows the comparisons of our ISR and ISP methods with other baselines in terms of accuracy. We do the evaluation under the noise generated by two different distributions as discussed previously. Each column represents a dataset affected by Gaussian noise or Rician noise. Each row shows the results of one method. In most cases, both ISR and ISP outperform other baselines, showing that our proposed methods effectively improve the robustness of the trained classifier. ISP achieves the best performance with a descent gap in test accuracy compared to the worst performer (N2V-Training). 
When we compare the results under Gaussian noise and Rician noise, one can observe that Rician noise impacts the trained models more than Gaussian noise does. This is due to the Gaussian distribution producing zero-mean noise, while the Rician distribution produces non-zero mean noise. Out of all the methods, N2V-Training performs the worst with a significant degradation in accuracy under Rician noise. Recall our previous observation in Section \ref{sec:intervened_data} that reconstructed images from Rician noise have larger background noise than the original noisy images, so it is understandable that N2V-Training loses its performance significantly in this setting.

\section{Conclusion}
\label{sec:conclusion}
We have presented two effective and efficient training methods, Influence-based Sample Reweighing and Influence-based Sample Perturbation, to improve the robustness of a brain tumor classification model against noisy training MR image data. Our comprehensive experiments showed that models trained with our methods are stable and robust against the effect of different noise distributions (Gaussian and Rician). Further, we showed that our methods could be directly applied to noisy training data without the need for additional clean images or denoising steps. Futher, our method can be applicable to other types of medical images such as X-ray or CT images. In future work, we will extend the ideas to more complex models (such as semi-supervised or unsupervised models), which are challenging but useful tasks in the medical imaging field since labeled training image data is limited. 

\section*{Acknowledgements}
This work was supported  in part by National Science Foundation under awards 1946391, the National Institute of General Medical Sciences of National Institutes of Health under award P20GM139768, and the Arkansas Integrative Metabolic Research Center at University of Arkansas.


\end{document}